\pdfoutput=1
\documentclass[11pt]{article}

\usepackage[final]{acl}

\usepackage{times}
\usepackage{latexsym}

\usepackage[T1]{fontenc}
\usepackage[utf8]{inputenc}

\usepackage{microtype}

\usepackage{inconsolata}

\usepackage{enumerate}
\usepackage{amsthm}
\usepackage{graphicx}
\usepackage{epstopdf}
\usepackage{caption}
\usepackage{color}
\usepackage{enumitem}
\usepackage{array}
\usepackage{tabularx}
\usepackage{multirow}
\usepackage{bm}
\usepackage{booktabs}

\usepackage{tcolorbox}
\tcbuselibrary{breakable}
\tcbuselibrary{skins}
\usepackage{caption}
\usepackage{subcaption}
\usepackage{listings}
\usepackage{wrapfig}
\usepackage{bbding}
\tcbuselibrary{skins}
\usepackage{colortbl}
\usepackage{fontawesome}

\usepackage{authblk}
\usepackage{amsmath}
\usepackage{hyperref}
\usepackage{etoolbox}
\preto{\abstractkeywords}{\nolinenumbers} 

\title{Accelerate Parallelizable Reasoning via Parallel Decoding within One Sequence}

\author[a,b]{Yijiong Yu}
\affil[a]{Tsinghua University}
\affil[b]{OpenCSG}

\begin{document}
\maketitle
\begin{abstract}

Recent advances in reasoning models have demonstrated significant improvements in accuracy by employing detailed and comprehensive reasoning processes. However, generating these lengthy reasoning sequences is computationally expensive and time-consuming. To address this inefficiency, we leverage the inherent parallelizability of certain tasks to accelerate the reasoning process. Specifically, when multiple parallel reasoning steps exist, we decode multiple tokens per forward pass via a tree-like attention mask within a single sequence, avoiding additional memory usage. Experimental results show that our method achieves up to nearly 100\% speedup in decoding while basically maintaining the answer quality. 
Our code is available in \href{https://github.com/yuyijiong/parallel-decoding-in-one-sequence}{https://github.com/yuyijiong/parallel-decoding-in-one-sequence}.

\end{abstract}

\begin{figure*}[htb]
    \centering
    \includegraphics[width=1\linewidth]{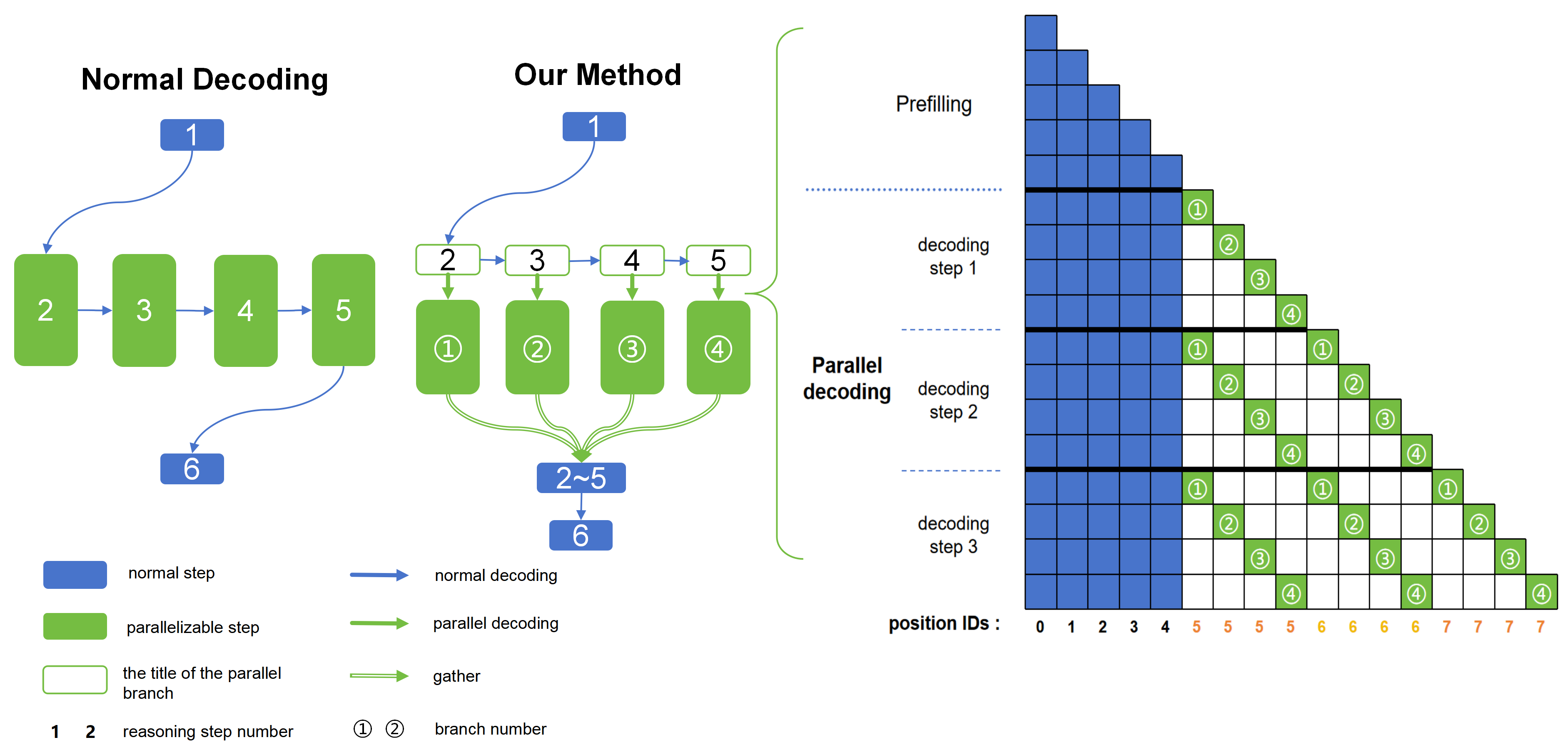}
    \caption{Comparison between our method and traditional decoding for a case with 6 steps and 4 of them are parallelizable. Our method generates only the title of the parallelizable steps sequentially, and then generates the remaining part via parallel decoding, using a special attention mask. The white blocks in the attention mask indicate "cannot see."}
    \label{fig:method}
\end{figure*}

\section{Introduction}

Large Language Models (LLMs), capable of performing complex reasoning processes, excel across a diverse array of tasks. However, their autoregressive decoding structure renders them inefficient for parallelizable reasoning tasks. Parallelizable reasoning tasks are those that, while requiring multiple steps, involve many independent steps, i.e., steps that can be executed concurrently due to the absence of a priority or causal relationship. For example, checking many items in a list or comparing multiple aspects of two products. However, in the decoding phase, LLMs must generate tokens sequentially, hindering the full utilization of the hardware's parallel computing capabilities.

Many previous works have used ``parallel decoding'' to accelerate LLM inference and increase hardware utilization. For example, skeleton-of-thoughts \cite{ning_skeleton--thought_2024} prompts the model to first generate a skeleton, and complete all the points in the skeleton in parallel via batch decoding or multiple API calls. MEDUSA \cite{cai_medusa_2024} trains multiple MEDUSA heads to respectively predict the next $n$-th token, and generate multiple candidates by combining the top predictions of each head, then it uses a tree attention mask to verify multiple candidates in parallel.

However, existing methods suffer from notable drawbacks. For skeleton-of-thoughts \cite{ning_skeleton--thought_2024}, firstly, it relies on batch decoding or multiple API calls, which exponentially increases the memory usage and API cost, forcing a reduction in batch size when GPU memory is limited. Secondly, it requires distinct prompt templates for different generation stages, preventing reuse of the KV cache. Lastly, it treats every point in their reasoning skeleton as independent, potentially neglecting causal relationships between different points. For MEDUSA \cite{cai_medusa_2024}, it must retrain multiple language modeling heads for a new model, which is not flexible and convenient enough.

To overcome these limitations, we propose a novel decoding method called "Parallel Decoding in One Sequence," specifically designed to address parallelizable tasks in LLM reasoning. This method operates in three stages: (1) identifying parallelizable steps in the reasoning process, (2) parallel decoding of these steps, and (3) concatenating the results and continuing generation. 

In stage 1, the model generates the reasoning process as normal, but is prompted to use a special token is to mark the beginning of each parallelizable step, only the title of which is generated, while the remaining part is omitted with an ellipse. So the number of output tokens is relatively few. In stage 2, we decode each step's subsequent tokens in parallel via a tree-like attention mask, where each step is treated as a branch, sharing the non-parallelizable tokens and having its own title tokens generated in stage 1 independently.
So the model can generate one token for every branch simultaneously in a single forward pass, significantly accelerating the generation. Finally, in stage 3, we concatenate the content generated in stage 2 of each parallel step and append them to the non-parallelizable part of the sequence, allowing the model to resume its reasoning process. The overall process is illustrated in Figure \ref{fig:method}.

Experiments on 3 parallelizable tasks demonstrate that our method significantly enhances decoding speed, with only a minor impact on generation quality. Although our method only targets some specific scenarios, it eliminates the need for additional memory usage and KV cache recomputation, as we do not create multiple sequences, making it particularly suitable for resource-constrained scenarios. Moreover, our method relies on the LLM itself to identify parallelizable steps but not assumes in advance that all steps are independent, avoiding the information loss caused by the inappropriately processing the steps with causal relationships in parallel. What is more, it does not require training or extra data, enabling greater flexibility and adaptability.

\section{Related Works}

The computational demands of LLMs have inspired the development of various techniques aimed at accelerating the decoding process. One widely recognized method is speculative decoding \cite{leviathan_fast_2023}, which employs a smaller assistant model to generate tokens first, allowing the main model to verify them—a process that leverages parallelizable verification for faster inference. Jacobi decoding \cite{santilli_accelerating_2023} constructs a batch of initial sequences using pad tokens, enabling iterative updates to parallelize token generation within each sequence. Medusa \cite{cai_medusa_2024} enhances the target LLM with auxiliary guess heads to facilitate self-speculation, achieving up to a threefold speedup on various tasks. Skeleton-of-thoughts \cite{ning_skeleton--thought_2024} prompts LLMs to generate a skeleton containing multiple points and completes each point in parallel by employing batch decoding or multiple API calls.

\section{Method}
Our method includes 3 stages: branch title generation, parallel decoding, concatenating and continuation. 

Stage 1.
Stage 1 is actually a prompt-guided special format generation. First, as normal, we give the model the task description. Then, we append an additional prompt (detailed in Appendix~\ref{app:prompt}), by which the model is prompted to mark parallelizable steps with \texttt{\#\#\#\#}, generating only titles followed by a colone and an ellipses (e.g., "\texttt{\#\#\#\# Step 1: ......}"). To ensure the correct format is generated, we manually manipulate the logits to force it to immediately generate an ellipsis after a colone, and a \texttt{\#\#\#\#} after an ellipsis. We group this bunch of consecutive parallel steps as a "parallel block", and we prompt it that a terminator (\texttt{\%\%\%\%}) should be generated to signify the end of the parallel block. This yields a compact skeleton while ensuring correct parallelism detection. The specific prompt we use in this stage is in Appendix \ref{app:prompt}.

Stage 2. 
We use each parallel step's title generated in stage 1 as each branch's prefix tokens. We signify the number of parallel steps as $n$. For $n$ parallel branches:
\begin{itemize}
    \item We reuse the KV cache of non-parallel steps (the part before the parallel block).
    \item We process $n$ tokens per forward pass using a \textit{tree-like} attention mask (as shown in Figure \ref{fig:method}), which ensures that the branches are isolated from one another while sharing non-parallel tokens.
    \item The position ids of the tokens processed in one decoding step are all the same. The position ID is incremented by one after each decoding step.
    \item We terminate the branch when \texttt{\#\#\#\#} is generated in this branch, and then pad it with padding tokens until all the other branches are terminated.
\end{itemize}

Stage 3.
We sequentially concatenate the fully decoded content of parallel steps in Stage 2 with the non-parallel part of the sequence. The model then resumes decoding as normal, and the KV cache of the non-parallel part can be reused. If "\#\#\#\#" is detected again, the process loops back to Stage 1, initiating the next parallel block.

\section{Experiments}

% \begin{table*}[h]
% \centering

% \begin{tabular}{lcccc}
% \toprule
% Method & Answer Quality & Decoding Speed (tokens/s) & Time (s) & Memory Usage (MB) \\
% \midrule
% Normal & 1.00 & 36.17 &9.36 & 17816 \\
% Ours & 1.00 & 61.09 &5.52& 17816 \\
% SoT & 0.43 & 29.24 &3.13& 19177 \\
% CoD & 1.00 & 36.51 &4.52& 17816 \\
% CoD + Ours & 1.00 & 42.58 &4.04& 17816 \\
% \bottomrule
% \end{tabular}
% \caption{Evaluation on the retrieval task.}
% \label{tab:retrieval}
% \end{table*}

% \begin{table*}[h]
% \centering
% \begin{tabular}{lcccc}
% \toprule
% Method & Answer Quality & Decoding Speed (tokens/s)& Time (s) & Memory Usage (MB) \\
% \midrule
% Normal & 2.77 & 36.25 &7.03& 22920 \\
% Ours & 2.88 & 57.68 &5.04& 22921 \\
% SoT & 2.88 & 13.72 &3.24& 40824 \\
% CoD & 2.33 & 35.27 &2.05& 22920 \\
% CoD + Ours & 1.92 & 27.69 &1.86& 22921 \\
% \bottomrule
% \end{tabular}
% \caption{Evaluation on the multi-document QA task.}
% \label{tab:QA}
% \end{table*}

% \begin{table*}[h]
% \centering
% \begin{tabular}{lcccc}
% \toprule
% Method & Answer Quality & Decoding Speed (tokens/s)& Time (s) & Memory Usage (MB) \\
% \midrule
% Normal & 3.83 & 36.25 &17.12& 17086 \\
% Ours & 3.32 & 54.72 &13.43& 17203 \\
% SoT & 3.18 & 85.85 & 5.12&18175 \\
% CoD & 2.93 & 36.50 &9.36& 17086 \\
% CoD + Ours & 2.37 & 33.22 &7.31& 17203 \\
% \bottomrule
% \end{tabular}
% \caption{Evaluation on the planning task.}
% \label{tab:planning}
% \end{table*}

\begin{table*}[h]
\centering
\begin{tabular}{l l c c c c}
\toprule
Task & Method & Answer Quality & Speed (tokens/s) & Time (s) & Memory (MB) \\
\midrule
\multirow{5}{*}{Retrieval}
& Normal & 1.00 & 36.17 & 9.36 & 17816 \\
& Ours & 1.00 & 61.09 & 5.52 & 17816 \\
& SoT & 0.43 & 29.24 & 3.13 & 19177 \\
& CoD & 1.00 & 36.51 & 4.52 & 17816 \\
& CoD + Ours & 1.00 & 42.58 & 4.04 & 17816 \\
\cmidrule(lr){1-6}
\multirow{5}{*}{Multi-Doc QA}
& Normal & 2.77 & 36.25 & 7.03 & 22920 \\
& Ours & 2.88 & 57.68 & 5.04 & 22921 \\
& SoT & 2.88 & 13.72 & 3.24 & 40824 \\
& CoD & 2.33 & 35.27 & 2.05 & 22920 \\
& CoD + Ours & 1.92 & 27.69 & 1.86 & 22921 \\
\cmidrule(lr){1-6}
\multirow{5}{*}{Planning}
& Normal & 3.83 & 36.25 & 17.12 & 17086 \\
& Ours & 3.32 & 54.72 & 13.43 & 17203 \\
& SoT & 3.18 & 85.85 & 5.12 & 18175 \\
& CoD & 2.93 & 36.50 & 9.36 & 17086 \\
& CoD + Ours & 2.37 & 33.22 & 7.31 & 17203 \\
\bottomrule
\end{tabular}
\caption{Evaluation results across 3 tasks: Retrieval, Multi-Document Question Answering, and Planning. We record the average answer quality (or accuracy), decoding speed, inference time, and GPU memory usage.}
\label{tab:combined}
\end{table*}

\subsection{Implementation Details}
The key innovation of our method lies in the modified causal attention mask, termed the "tree-like mask." While FlashAttention-2 \cite{dao2023flashattention2fasterattentionbetter} accelerates attention computation and saves memory, it does not support custom masks. Thus, we modified its source code to track two additional parameters: the number of branches ($n$) and the position where parallel decoding begins. They both derived from content generated in Stage 1. Specifically, where the first special mark "\#\#\#\#" appears determines the position where parallel decoding begins, and the number of special marks "\#\#\#\#" determines the number of branches.

For experimentation, we use Qwen2.5 \cite{team_qwen25_2024} implemented via the HuggingFace Transformers \cite{wolf-etal-2020-transformers}, coupled with our customized FlashAttention-2 package. All models are run on an single A100 GPU with bfloat16 precision. We set temperature to 0 for all the generation experiments.

To make a baseline, we provide the model with the task description and the additional prompt used in stage 1 of our method to ensure that the length and format of the answers are basically consistent, and let it generate answers with standard decoding methods. 

\subsection{Data and Evaluation}
Our method targets parallelizable tasks, so we select a retrieval task, a multi-document QA task and a planning task, which all require reasoning and involve parallelizable steps, such as the analysis of individual items (e.g., multiple students, documents or aspects). The retrieval task and the multi-document QA task belongs to long-context tasks, with a long reference text in the prompt on which the model must answer based on. While for the planning task, our prompt is relatively short, letting the model answer based on its internal knowledge. The examples of these 3 tasks are shown in \ref{app:example}.

For the retrieval task, we use the student resume retrieval dataset from difficult-retrieval \cite{yu_hyper-multi-step_2024}, containing 100 samples per setting. The model need to retrieve the student whose GPA falls within a specified range among 10 students, requiring parallel GPA analysis. The average prompt length is 680.

The multi-document QA task involves "2WikiMultihopQA" from LongBench \cite{bai_longbench_2023}, comprising 200 samples. The model must answer questions based on 10 documents, with only one of them containing relevant information, necessitating parallel document analysis. The average prompt length is 5343.

The planning task requires the model to first analyze multiple aspects and summarize them to give a final plan. For example, "Please analyze the current application status of artificial intelligence from three dimensions: technological challenges, ethical risks, and future development trends. Finally, provide an overall evaluation." Obviously, different aspects can be analyzed in parallel. We design 100 samples with reference answers using QwQ-32B \cite{qwq32b}, which covers topics from various fields such as technology, politics, and business. The number of branches for the task ranges from 2 to 10, with an average of 4.4 branches, and the average prompt length is 99.

We use exact-match to assess the retrieval task and use GPT-4o-2024-11-20 \cite{openai_gpt-4o_2024} to rate (giving a score from 1 to 5) the answer quality of the QA task and planning task based on the reference answer. The prompt for GPT-4o rating is shown in Appendix \ref{app:gpt rating}. For each sample, we recorded the total time (including both prefilling and decoding) it took for the model to complete the inference, and the decoding speed, measured in average number of tokens generated per second during the decoding phase.

For comparison, we also conduct experiments on Skeleton-of-thoughts (SoT) \cite{ning_skeleton--thought_2024} using the same data. We use the "batch\_outline" mode, i.e. batch decoding, provided by its source code to generate the responses. What is more, we compare our method with Chain-of-Draft (CoD) \cite{xu_chain_2025}, a prompting strategy encouraging the model to output more concise reasoning steps to accelerate reasoning. We also try combining CoD with our method to achieve a higher acceleration. The models used in all the methods are Qwen2.5-7B-Instruct \cite{team_qwen25_2024}.

We also experiment with other models, such as Qwen2.5-14B-Instruct \cite{team_qwen25_2024}, phi-4-mini-instruct \cite{abdin2024phi4technicalreport}, and QwQ-32B \cite{qwq32b}.

\subsection{Results}
Table \ref{tab:combined} presents the results of the 3 tasks. For retrieval, Our method nearly doubles decoding speed for both models, and the accuracy is not decreased, which means our method suits the retrieval task very well. We show an example of the student retrieval task using our method in Appendix \ref{app:example}. 

For multi-document QA, our method improves the decoding speed by nearly 60\%, and the answer quality even increase a little, which is within the normal fluctuation range, showing it suits QA task too. 

For the planning task, our method improves the decoding speed by over 50\%, and the answer quality is basically maintained, as evidenced by minor decrease in GPT-4o rating scores, representing an acceptable trade-off. 

In contrast, although SoT \cite{ning_skeleton--thought_2024} takes less time in all the tasks than our method, it causes a significant increase on memory usage in multi-document QA because of the long context length of this task, and a more severe decline in answer quality in retrieval and planning. CoD \cite{xu_chain_2025} can indeed save time, and when combined with our method, it saves even more. However, it forces the model to output overly concise reasoning steps, inevitably leading to a severe answer quality decline in QA and planning.

\begin{table}[t]
	\centering
    \resizebox{\columnwidth}{!}{
	\begin{tabular}{lc|cc}
		\toprule
		Model &  Method & Answer quality   & Speed (tokens/s)\\
            \midrule
		\multirow{2}{*}{Qwen2.5-14b} & Normal & 3.56  & 21.1\\
		 & Ours & 3.75  & 35.8\\
                    \midrule
		\multirow{2}{*}{phi-4-mini} & Normal & 2.95  & 35.1\\
		 & Ours & 2.78   & 51.5\\
		\bottomrule
	\end{tabular}}
  \caption{Evaluation of other models on the planning task. }
	\label{tab:other model}
\end{table}

Experiments results (shown in Table \ref{tab:other model}) of other instruction-tuned models like Qwen2.5-14b-Instruct and phi-4-mini-instruct are similar to that of Qwen2.5-7b-Instruct, showing the generalizability of our method. However, we find our method can hardly work when using reasoning models such as QwQ-32B \cite{qwq32b}, because them always output a very long ``thinking'' process first, where the content does not follow any format instructions, making our method unable to function.

\section{Conclusion}
This paper introduces "Parallel Decoding in One Sequence," (PDOS) a novel method to accelerate the reasoning process of LLMs in parallelizable tasks, through prompting and attention mask modification, which substantially reduces inference time while basically maintaining the model's flexibility and quality. Critically, it achieves these advantages without additional memory usage or recomputing the KV caches of the prefilling stage. Future work could explore extending this approach to more complex and diverse task types or investigating its applicability across different model architectures and sizes.

\section{Limitations}
For small models or large reasoning models, Stage 1 may fail to produce the required format, preventing subsequent stages from functioning, because a non-standard format impedes recognition of branch titles and the determination of the number of parallel branches.

Furthermore, our method has been tested only on synthetic datasets. For more complex tasks, where the reasoning processes may exhibit greater uncertainty, extending our approach may present challenges.

\bibliography{custom2}

\begin{thebibliography}{13}
\providecommand{\natexlab}[1]{#1}

\bibitem[{Abdin et~al.(2024)Abdin, Aneja, Behl, Bubeck, Eldan, Gunasekar, Harrison, Hewett, Javaheripi, Kauffmann, Lee, Lee, Li, Liu, Mendes, Nguyen, Price, de~Rosa, Saarikivi, Salim, Shah, Wang, Ward, Wu, Yu, Zhang, and Zhang}]{abdin2024phi4technicalreport}
Marah Abdin, Jyoti Aneja, Harkirat Behl, Sébastien Bubeck, Ronen Eldan, Suriya Gunasekar, Michael Harrison, Russell~J. Hewett, Mojan Javaheripi, Piero Kauffmann, James~R. Lee, Yin~Tat Lee, Yuanzhi Li, Weishung Liu, Caio C.~T. Mendes, Anh Nguyen, Eric Price, Gustavo de~Rosa, Olli Saarikivi, Adil Salim, Shital Shah, Xin Wang, Rachel Ward, Yue Wu, Dingli Yu, Cyril Zhang, and Yi~Zhang. 2024.
\newblock \href {https://arxiv.org/abs/2412.08905} {Phi-4 technical report}.
\newblock \emph{Preprint}, arXiv:2412.08905.

\bibitem[{Bai et~al.(2023)Bai, Lv, Zhang, Lyu, Tang, Huang, Du, Liu, Zeng, Hou, Dong, Tang, and Li}]{bai_longbench_2023}
Yushi Bai, Xin Lv, Jiajie Zhang, Hongchang Lyu, Jiankai Tang, Zhidian Huang, Zhengxiao Du, Xiao Liu, Aohan Zeng, Lei Hou, Yuxiao Dong, Jie Tang, and Juanzi Li. 2023.
\newblock \href {https://arxiv.org/abs/2308.14508} {{LongBench}: {A} {Bilingual}, {Multitask} {Benchmark} for {Long} {Context} {Understanding}}.

\bibitem[{Cai et~al.(2024)Cai, Li, Geng, Peng, Lee, Chen, and Dao}]{cai_medusa_2024}
Tianle Cai, Yuhong Li, Zhengyang Geng, Hongwu Peng, Jason~D. Lee, Deming Chen, and Tri Dao. 2024.
\newblock \href {https://doi.org/10.48550/arXiv.2401.10774} {Medusa: Simple {LLM} inference acceleration framework with multiple decoding heads}.
\newblock \emph{Preprint}, arxiv:2401.10774 [cs].

\bibitem[{Dao(2023)}]{dao2023flashattention2fasterattentionbetter}
Tri Dao. 2023.
\newblock \href {https://arxiv.org/abs/2307.08691} {Flashattention-2: Faster attention with better parallelism and work partitioning}.
\newblock \emph{Preprint}, arXiv:2307.08691.

\bibitem[{Leviathan et~al.(2023)Leviathan, Kalman, and Matias}]{leviathan_fast_2023}
Yaniv Leviathan, Matan Kalman, and Yossi Matias. 2023.
\newblock \href {https://doi.org/10.48550/arXiv.2211.17192} {Fast inference from transformers via speculative decoding}.
\newblock \emph{Preprint}, arxiv:2211.17192 [cs].

\bibitem[{Ning et~al.(2024)Ning, Lin, Zhou, Wang, Yang, and Wang}]{ning_skeleton--thought_2024}
Xuefei Ning, Zinan Lin, Zixuan Zhou, Zifu Wang, Huazhong Yang, and Yu~Wang. 2024.
\newblock \href {https://doi.org/10.48550/arXiv.2307.15337} {Skeleton-of-thought: Prompting {LLMs} for efficient parallel generation}.
\newblock \emph{Preprint}, arxiv:2307.15337 [cs].

\bibitem[{OpenAI(2024)}]{openai_gpt-4o_2024}
OpenAI. 2024.
\newblock \href {https://arxiv.org/abs/2410.21276} {{GPT}-4o {System} {Card}}.

\bibitem[{Santilli et~al.(2023)Santilli, Severino, Postolache, Maiorca, Mancusi, Marin, and Rodolà}]{santilli_accelerating_2023}
Andrea Santilli, Silvio Severino, Emilian Postolache, Valentino Maiorca, Michele Mancusi, Riccardo Marin, and Emanuele Rodolà. 2023.
\newblock \href {https://doi.org/10.18653/v1/2023.acl-long.689} {Accelerating transformer inference for translation via parallel decoding}.
\newblock In \emph{Proceedings of the 61st Annual Meeting of the Association for Computational Linguistics (Volume 1: Long Papers)}, pages 12336--12355.

\bibitem[{Team(2024)}]{team_qwen25_2024}
Qwen Team. 2024.
\newblock \href {http://qwenlm.github.io/blog/qwen2.5/} {Qwen2.5: A party of foundation models!}
\newblock Section: blog.

\bibitem[{Team(2025)}]{qwq32b}
Qwen Team. 2025.
\newblock \href {https://qwenlm.github.io/blog/qwq-32b/} {Qwq-32b: Embracing the power of reinforcement learning}.

\bibitem[{Wolf et~al.(2020)Wolf, Debut, Sanh, Chaumond, Delangue, Moi, Cistac, Rault, Louf, Funtowicz, Davison, Shleifer, von Platen, Ma, Jernite, Plu, Xu, Scao, Gugger, Drame, Lhoest, and Rush}]{wolf-etal-2020-transformers}
Thomas Wolf, Lysandre Debut, Victor Sanh, Julien Chaumond, Clement Delangue, Anthony Moi, Pierric Cistac, Tim Rault, Rémi Louf, Morgan Funtowicz, Joe Davison, Sam Shleifer, Patrick von Platen, Clara Ma, Yacine Jernite, Julien Plu, Canwen Xu, Teven~Le Scao, Sylvain Gugger, Mariama Drame, Quentin Lhoest, and Alexander~M. Rush. 2020.
\newblock \href {https://www.aclweb.org/anthology/2020.emnlp-demos.6} {Transformers: State-of-the-art natural language processing}.
\newblock In \emph{Proceedings of the 2020 Conference on Empirical Methods in Natural Language Processing: System Demonstrations}, pages 38--45, Online. Association for Computational Linguistics.

\bibitem[{Xu et~al.(2025)Xu, Xie, Zhao, and He}]{xu_chain_2025}
Silei Xu, Wenhao Xie, Lingxiao Zhao, and Pengcheng He. 2025.
\newblock \href {https://doi.org/10.48550/arXiv.2502.18600} {Chain of draft: Thinking faster by writing less}.
\newblock \emph{Preprint}, arxiv:2502.18600 [cs].

\bibitem[{Yu et~al.(2024)Yu, Xiufa, Jianwei, Xu, Guangyao, Jiancheng, Huang, Qi, Wang, Liu, Chen, and Pei}]{yu_hyper-multi-step_2024}
Yijiong Yu, Ma~Xiufa, Fang Jianwei, Zhi Xu, Su~Guangyao, Wang Jiancheng, Yongfeng Huang, Zhixiao Qi, Wei Wang, Weifeng Liu, Ran Chen, and Ji~Pei. 2024.
\newblock \href {https://arxiv.org/abs/2410.04422} {Hyper-multi-step: {The} {Truth} {Behind} {Difficult} {Long}-context {Tasks}}.

\end{thebibliography}
%\clearpage
\appendix

\section{Prompt}
\label{app:prompt}

The additional prompt we use in stage 1 to let the model generate in a skeleton-like format is:
\begin{tcolorbox}[colback=white,breakable]
When you need to sequentially handle multiple parallel steps (the steps are individual, for example, analyzing multiple individual documents, planning multiple branches, evaluating multiple aspects) during the reasoning process, you must strictly adhere to the following format: You need to prefix each step with '\#\#\#\#', followed by the step's title, and then a colon ':' (an English colon). \\
After all the steps are completed, you need to output '\#\#\#\#\%\%\%\%', and only then can you proceed with the subsequent reasoning process.\\

Example 1:\\
Question: [Resumes of A, B, C, D] Please analyze which of the four individuals best meets the requirements.\\
Answer: Let us analyze the strengths of each person based on their resumes.\\
\#\#\#\#Strengths of A:......\\
\#\#\#\#Strengths of B:......\\
\#\#\#\#Strengths of C:......\\
\#\#\#\#Strengths of D:...... \\
\#\#\#\#\%\%\%\%
Therefore, I believe that A's resume best meets the requirements.'\\

Example 2:\\
Question: [document 1,2,3,4] Based on the information in the documents, what is the birthday of Jack?\\
Answer: Let us analyze each documents.\\
\#\#\#\#document 1:......\\
\#\#\#\#document 2:......\\
\#\#\#\#document 3:......\\
\#\#\#\#document 4:...... \\
\#\#\#\#\%\%\%\%
Therefore, Jack's birthday is 5th, May, 2000.'\\

Example 3:\\
Question: Please analyze the development status of China from the aspects of economy, politics, culture, and society.\\
Answer: Let us analyze from four aspects.\\
\#\#\#\#Economy:......\\
\#\#\#\#Politics:......\\
\#\#\#\#Culture:......\\
\#\#\#\#Society:...... \\
\#\#\#\#\%\%\%\%
Therefore, we can conclude that the development status of China is......\\

Note that the examples is only used to illustrate the format; the specific content should closely revolve around the specific task requirements.\\
If you see that the part after the colon in a certain step is replaced with an ellipsis, it means that the specific content does not need to be provided for this reasoning step, only the title is required, and you should directly proceed to the next parallel step. Otherwise, give a complete and clear analysis for this step.\\
Please be careful, do not forget any necessary steps. \\
Note that only a branch step should start with \#\#\#\#. If it is a stem or general step, you should not add \#\#\#\#.
\end{tcolorbox}

\section{Examples of the Prompt of the 3 Tasks and the Model's Output}
\label{app:example}

Here we show an example of what the model generated when using our parallel decoding method, in the student retrieval task. We omit some content in the prompt. As we can see, in stage 1, the model only generate the skeleton, where the specific analysis process of each branch is represented by ellipses. Then in stage 2, each branch's specific reasoning process is completed by parallel decoding. Finally, in stage 3, all the reasoning processes are concatenated, and the model generate the final conclusion.

\begin{tcolorbox}[colback=white,breakable]

\textbf{Prompt:}\\

Here are 10 students' resumes:\\

The student named Nitin Sharma is 29 years old, graduated from IIM Ahmedabad with a GPA of 4.09. He/She is interested in Finance, Cooking, Travel and his/her self-introduction is: Finance enthusiast who loves cooking and exploring diverse cuisines.\\
......\\
The student named Lily Wilson is 23 years old, graduated from University of Brighton with a GPA of 2.45. He/She is interested in Political Science, Activism, Public Speaking and his/her self-introduction is: I am dedicated to activism and often channel my thoughts into public speaking.\\
......\\

Question: Which student has a GPA between 2.36 and 2.75? Please give your final answer (the student's name) in the format of "name: {answer}" \\
You should check every student to judge whether he meets the requirement in your reasoning process.
\tcbline
\textbf{Stage 1:}\\

Let us analyze each student's GPA to find the one between 2.36 and 2.75.

\#\#\#\#Nitin Sharma: ......

\#\#\#\#Hallie Turner: ......

\#\#\#\#Hikaru Sawai: ......

\#\#\#\#Abigail Clark: ......

\#\#\#\#Charlotte Phillips: ......

\#\#\#\#Lily Wilson: ......

\#\#\#\#Cao Ling: ......

\#\#\#\#William Stewart: ......

\#\#\#\#Sienna Torres: ......

\#\#\#\#Eli White: ......
\#\#\#\#\%\%\%\%

\tcbline
\textbf{Stage 2:}\\

\#\#\#\#Nitin Sharma: GPA is 4.09, which is not between 2.36 and 2.75.

\#\#\#\#Hallie Turner: GPA is 3.68, which is not between 2.36 and 2.75.

\#\#\#\#Hikaru Sawai: GPA is 2.98, which is higher than 2.75.

\#\#\#\#Abigail Clark: GPA is 1.83, which is below the range.

\#\#\#\#Charlotte Phillips: GPA is 1.54, which is below the range.

\#\#\#\#Lily Wilson: GPA is 2.45, which falls within the range of 2.36 to 2.75.

\#\#\#\#Cao Ling: Cao Ling's GPA is 2.82, which is above 2.75.

\#\#\#\#William Stewart: GPA is 4.04, which is not between 2.36 and 2.75.

\#\#\#\#Sienna Torres: GPA is 4.48, which is not between 2.36 and 2.75.

\#\#\#\#Eli White: GPA is 4.61, which is not between 2.36 and 2.75.

\tcbline
\textbf{Stage 3:}\\

Based on the analysis, Lily Wilson is the student whose GPA falls within the specified range. 

Name: Lily Wilson
\end{tcolorbox}

An example of the prompt of the multi-document QA task:
\begin{tcolorbox}[colback=white,breakable]
Passage 1:
Waldrada of Lotharingia. Waldrada was the mistress, and later the wife, of Lothair II of Lotharingia.
......\\

Passage 2: ......\\

Passage 3: ......\\

Question: Where was the wife of Francis I Rákóczi born?\\

You should check each document one by one (no matter whether it is relevant), and analyze its content to judge whether it provides information about the question. After analysis, output your final answer in the format of 'Answer: your concise answer.

\end{tcolorbox}

An example of the prompt of the planning task:
\begin{tcolorbox}[colback=white,breakable]

Please analyze the five core challenges that drone delivery may face in practical applications and propose at least one solution for each challenge. 

Require analysis from five different dimensions: technical feasibility, regulatory limitations, social acceptance, business model, and environmental impact, and ultimately form a comprehensive evaluation report.

\end{tcolorbox}

\section{Prompt for GPT Rating}
\label{app:gpt rating}

The prompt used for GPT-4o to rate the answer quality is as follows. For rating, we set the temperature to 0 and max generated tokens to 4k.

\begin{tcolorbox}[colback=white,breakable]

You are an assistant skilled in evaluating the quality of responses.\\
Please act as an impartial judge to assess the accuracy, usefulness, and clarity of an AI assistant's response to a user's question. \\
I will provide the question, reference answer, and the AI assistant's response. \\
You need to evaluate the quality of the AI assistant's response and give a score from 1 to 5.\\

1 point: The response is completely incorrect, unhelpful to the question, or poorly formatted with low readability.\\
2 points: The response is basically incorrect, somewhat helpful to the question, or somewhat poorly formatted with low readability.\\
3 points: The response is basically correct, moderately helpful to the question, or moderately formatted with average readability.\\
4 points: The response is correct, helpful to the question, or well-formatted with good readability.\\
5 points: The response is completely correct, highly helpful to the question, and neatly formatted with excellent readability.\\

Please provide your evaluation in the format: "Reason: ... Score: ...".\\

User's question: \{question\}\\
Reference answer: \{reference\}\\
AI assistant's answer: \{answer\}\\

\end{tcolorbox}

% \section{Experiments on other models}
% \label{app:other model}
% \input{results_other_model}

\end{document}